\documentclass[10pt,letterpaper]{article}

\usepackage{cogsci}

\cogscifinalcopy 

\usepackage{pslatex}
\usepackage{apacite}
\usepackage{float} 
\usepackage{graphicx,xcolor}
\usepackage{amsmath}
\usepackage{booktabs}
\usepackage{amssymb}
\usepackage{supertabular}

\title{
Do Neural Language Representations Learn Physical Commonsense?} 
 
\author{
\\ \\
 {\large \bf Maxwell Forbes$\dagger$, Ari Holtzman$\dagger\ddagger$, and Yejin Choi$\dagger\ddagger$ } \\ 
  \{mbforbes, ahai, yejin\}@cs.washington.edu\\
  $\dagger$Paul G. Allen School of Computer Science and Engineering, University of Washington\\  
  $\ddagger$Allen Institute for Artificial Intelligence
}

\begin{document}

\setlength{\abovedisplayskip}{2pt}
\setlength{\belowdisplayskip}{2pt}

\maketitle

\begin{abstract}

Humans understand language based on rich background knowledge about how the physical world works, which in turn allows us to reason about the physical world through language.  
In addition to the \emph{properties} of objects (e.g., \textit{boats require fuel}) and their \emph{affordances}, i.e., the actions that are applicable to them (e.g., \textit{boats can be driven}), we can also reason about \emph{if--then} inferences between what properties of objects imply the kind of actions that are applicable to them (e.g., that \textit{if we can drive something then it likely requires fuel}).

In this paper, we investigate the extent to which state-of-the-art neural language representations, trained on a vast amount of natural language text, demonstrate physical commonsense reasoning. 
While recent advancements of neural language models have demonstrated strong performance on various types of natural language inference tasks, our study based on a dataset of over 200k newly collected annotations suggests that neural language representations still only learn associations that are explicitly written down.\footnote{Visit \url{https://mbforbes.github.io/physical-commonsense} for our data, code, and more project information.}

\end{abstract}

\section{Introduction}

\begin{figure}[t!]

\begin{center}
\includegraphics[width=0.99\linewidth]{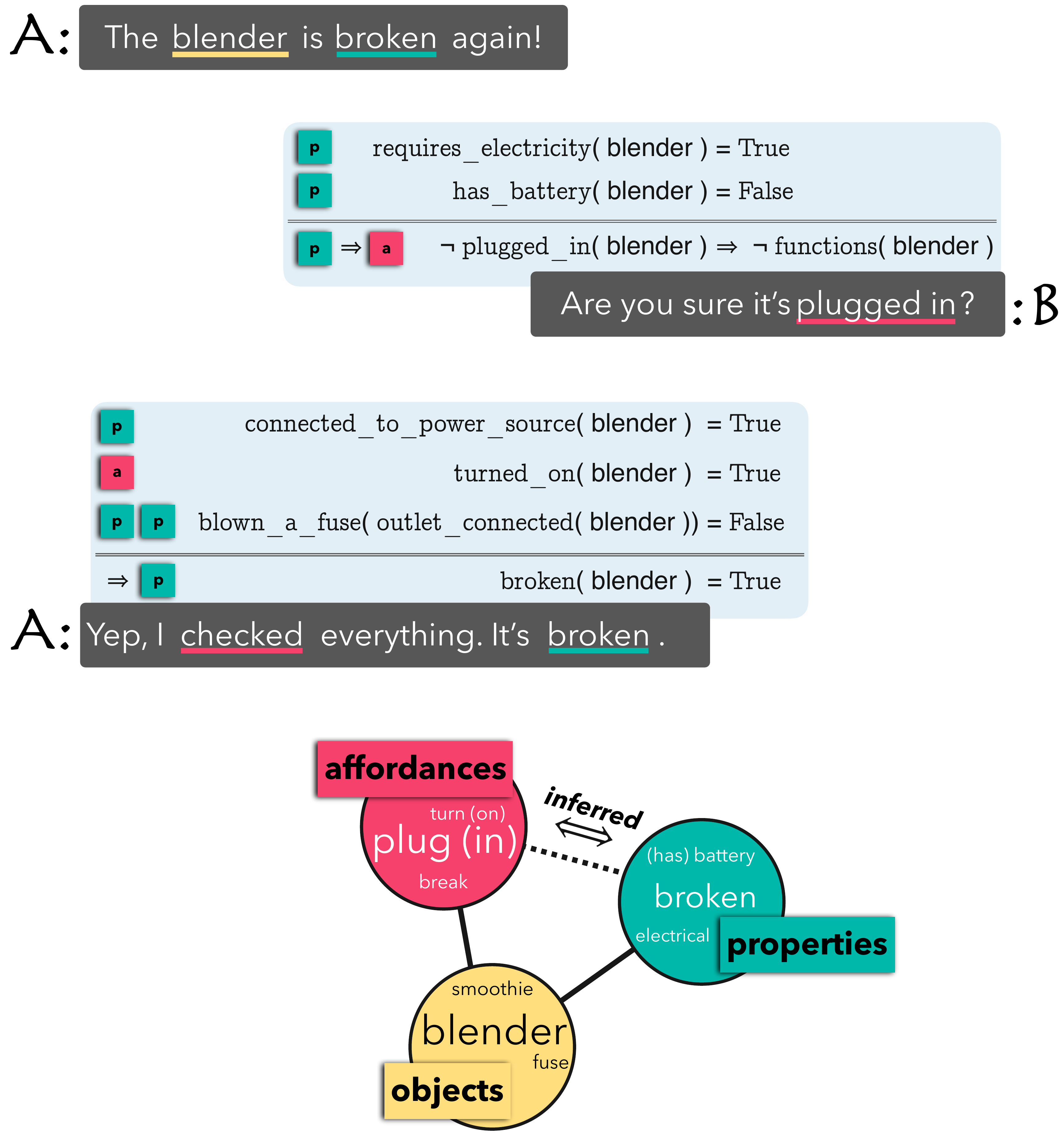}
\end{center}

\caption{Natural language communication often requires reasoning about  the affordances of objects (i.e., what actions are applicable to objects) from the properties of the objects (e.g., whether an object is edible, stationary, or requires electricity) and vice versa. We study the extent to which neural networks trained on a large amount of text can recover various aspects of physical commonsense knowledge.}
\label{fig:headlne}

\end{figure}

Understanding everyday natural language communication requires a rich spectrum of physical commonsense knowledge.
Consider the example dialog sketched in Figure~\ref{fig:headlne}.
A simple observation that, \textit{``The blender is broken again!''} triggers myriad pieces of implied understanding (e.g., that something which requires electricity will only work with a source of power).
Such knowledge is rarely stated explicitly \cite{van2010extracting}, and instead can be inferred on-the-fly as needed.

In this paper, we study physical commonsense knowledge underlying  natural language understanding, organized as interactions among three distinct concepts: (i) objects, (ii) their attributes (properties), and (iii) the actions that can be applied to them (affordances) (Figure~\ref{fig:headlne}, bottom). 
The premise of our study is that language models trained on a sufficiently large amount of text can recover a great deal of physical commonsense knowledge about each of these concepts. 
However, aspects of this knowledge may only be implicit in natural language utterances. 
For example, answering a question from the Winograd Schema Challenge \cite{Levesque:2012:WSC:3031843.3031909}---''The trophy would not \emph{fit} in the brown suitcase because it was too \emph{big}. What was too big?''---implicitly requires the physical commonsense reasoning that \emph{``in order to fit X in Y, X should be relatively smaller compared to Y,''} which essentially requires reasoning about the affordances of objects (\textit{fit X in Y}) from their attributes (relative sizes of X and Y). 

We investigate the extent to which neural language models trained on a massive amount of text demonstrate various aspects of physical commonsense knowledge and reasoning. 
Our analysis includes word embeddings such as GloVe \cite{pennington2014glove}, as well as more recent contextualized representations like ELMo \cite{peters2018deep} and BERT \cite{devlin2018bert}.
Such models are trained without supervision by exposing them to billions of words, and allowing them to extract patterns purely from token prediction tasks that can be derived directly from raw text.
These language representation models have established unprecedented performance on a wide range of evaluations, including natural language inference and commonsense reasoning.

How much do these large, unsupervised models of language learn about physical commonsense knowledge?
Some recent work has studied the capabilities of word embeddings to predict an object's properties \cite{rubinstein2015well,lucy2017distributional}.
Motivated by these efforts to understand language representations, we present several contributions.
We propose two datasets: the \textit{abstract} dataset, a refreshed version of the McRate dataset \cite{mcrae2005semantic}, pruned and densely annotated to eliminate false negatives present in previous work; and the \textit{situated dataset}, with annotations for objects' properties and affordances in real-world images sampled from the MS COCO dataset \cite{lin2014microsoft}.
As in previous work, we consider the prediction task of linking objects and their properties (O$\longleftrightarrow$P), but with our new situated dataset, we are also able to study the connection between objects and their affordances (O$\longleftrightarrow$A), as well as between affordances and properties (A$\longleftrightarrow$P).
We also study the latest models from the natural language processing community (ELMo, BERT) using in-context word representations, and present results for all of our proposed datasets and tasks.
Our analysis suggests that current neural language representations are proficient at guessing the affordances and properties of objects, but lack the ability to reason about the relationship between affordances and properties itself.
\section{Characterizing Objects through\\Properties and Affordances}

\subsection{Properties}

We use the term \textit{properties} to refer to the static characteristics of objects.
They encompass our commonsense understanding of what something is like.
For example, we might say that an \textit{apple} has the property of being \textit{edible}, or that a \textit{plant} is \textit{stationary}.

As with \citeA{mcrae2005semantic}, properties capture the general perception of a thing.
Exceptions naturally arise.
For example, specific instances can violate the general properties of an object, such as the inediblilty of a rotten apple. Additionally, subtypes can diverge from the exemplar of a category: the Venus flytrap is a plant with the ability to move.

\subsection{Affordances}

We express an object's actions with verbs.
One way to focus on understanding the actions of objects is to focus on their \textit{affordances}.
Coined by \citeA{gibson1966senses}, this term initially described animal-perceived uses for an object, but has since come to mean the perceived uses of an object in a given environment \cite{norman1988design,gaver1991technology}.

Here, we take a simpler, human-centric definition.
We consider an object's affordances to be, ``what actions do humans take with an object?''
For example, \textit{boots} commonly afford \textit{wear}, \textit{kick off}, \textit{lace up}, and \textit{put on}.

\subsection{Inference Between Affordances and Properties}

Affordances and properties exhibit a surprising connection.
As humans, we are able to infer many of an object's affordances based on its properties (A$\leftarrow$P).
The same is also true in the reverse (A$\rightarrow$P).

Consider an exchange: \textit{``You think you could fit that boulder in your truck?'' ``No way! That thing was so big you could go for a hike on it.''} We might sketch out some of this information as:

\begin{align*}
    \textit{fit } x \textit{ into } y &\implies x <^{\text{size}} y \\
    \text{hike}(\text{x}) &\implies x \gg^{\text{size}} \textsc{Human} \\
\end{align*}

While the above information only concerns a property's \textbf{\emph{relative}} value (comparative size), a broad range of information can be inferred between affordances and properties. Our focus in this work is on \textbf{\emph{absolute}} properties, for example:

\begin{align*}
    \textit{She \textbf{plugged in}} & \textit{ her robot.} \\
    \text{plug-in}(x) & \implies \text{uses-electricity}(x) \\
    \textit{She \textbf{looked through}} & \textit{ the keyhole.} \\
    \text{look-through}(x) & \implies \text{transparent}(x) \\
    \textit{He \textbf{poured} coffee} & \textit{ into the cup} \\
    \text{pour-into}(x) & \implies \text{holds-liquid}(x) \\
    \textit{It \textbf{shattered} on} & \textit{ the floor.} \\
    \text{shatter}(\text{x}) & \implies \text{rigid}(x) \\
\end{align*}

The implications ($\implies$) should be taken with a probabilistic grain of salt.
However, they capture our intuitions about what we expect to be true.
Wouldn't it be surprising to shatter something that isn't rigid, or plug-in something that doesn't take power?

Humans use the link between affordances and properties to recover information.
Can machine learning models do the same?
It is is difficult to model these implications based on text alone because there is no direct evidence for the implied information.
Any implication that can be trivially understood by a person is precisely the kind of information left unsaid. Who would write, \textit{``If I can walk inside my house, I know that my house is bigger than I am?''} Nevertheless, we naturally understand that: $x \textit{ walk-inside } y \implies x <^\text{size} y$.

Directly attacking the link between affordances and properties requires access to implications across the edges.
Without such information, we can use objects as a proxy to understand how much modern neural networks know about this edge.
For example, taking an object like \textit{boots}, and using only its top affordances \textit{wear}, \textit{kick off}, and \textit{lace up}, can we predict its properties?
\section{Experiments}

\begin{table}[t]
\centering

%
%

\begin{flushleft}\textbf{Statistics}\end{flushleft}
\resizebox{\linewidth}{!}{%
\begin{tabular}{@{}lrl@{}}
\toprule
           & \textbf{Total} & \textbf{Statistics}                           \\ \midrule
\textbf{Abstract}    &    & \\
Objects    & 514   & 411 train / 103 test                  \\
Properties & 50    & obj/prop: 60 median (3 min, 302 max) \\
           &       & prop/obj: 8 median (1 min, 23 max)   \\ \vspace*{4mm}
Annotations & 77,100 & 3 anns/datum \\

\textbf{Situated}    &    & \\
Objects    & 1,024   & 80 unique, split: 64 train / 16 test \\
Properties    & 50   & \\
Affordances    & 3072   & 3 affordances / object (by design)\\
Annotations & 156,672 & 3 anns/datum \\
\bottomrule
\end{tabular}%
}

\vspace*{6mm}

%
%

\begin{flushleft}\textbf{Examples}\end{flushleft}
\resizebox{\linewidth}{!}{%
\begin{tabular}{@{}lll@{}}
\toprule
\textbf{Objects}     & \textbf{Properties}  & \textbf{Affordances}       \\
\midrule
\textit{harmonica, van} & \textit{expensive, squishy} & \textit{pick up, remove} \\
\textit{potato, shovel} & \textit{used as a tool for cooking} & \textit{pet, talk to} \\
\textit{cat, bed} & \textit{decorative, fun} & \textit{cook, throw out} \\ \bottomrule
\end{tabular}%
}

%
%

\caption{Statistics and examples for the proposed abstract and situated datasets (based on \cite{mcrae2005semantic} and \cite{lin2014microsoft}).}
\label{table:dataset-stats}

\end{table}

\begin{table*}[]
\resizebox{\textwidth}{!}{%
\begin{tabular}{@{}llllllllllllllllllll@{}}
\toprule
 & \multicolumn{4}{c}{\textbf{Abstract}} &  & \multicolumn{14}{c}{\textbf{Situated}} \\ \midrule
 & \multicolumn{4}{c}{O $\longleftrightarrow$ P} &  & \multicolumn{4}{c}{O $\longleftrightarrow$ P} &  & \multicolumn{4}{c}{O $\longleftrightarrow$ A} &  & \multicolumn{4}{c}{A $\longleftrightarrow$ P} \\ \cmidrule(lr){2-5} \cmidrule(lr){7-10} \cmidrule(lr){12-15} \cmidrule(l){17-20} 
 & \textit{obj} & \textit{prop} & \textit{$\mu$F1} & \textit{sig} &  & \textit{obj} & \textit{prop} & \textit{$\mu$F1} & \textit{sig} &  & \textit{obj} & \textit{aff} & \textit{$\mu$F1} & \textit{sig} &  & \textit{aff} & \textit{prop} & \textit{$\mu$F1} & \textit{sig} \\ \midrule
\textsc{Random} & 0.25 & 0.26 & 0.26 & *** &  & 0.24 & 0.25 & 0.22 & *** &  & 0.53 & 0.62 & 0.51 & *** &  & 0.24 & \textbf{0.26} & 0.23 & *** \\
\textsc{Majority} & 0.34 & 0.11 & 0.31 & *** &  & 0.16 & 0.05 & 0.17 & *** &  & 0.82 & 0.68 & 0.82 & *** &  & 0.18 & 0.05 & 0.17 & *** \\
\textsc{GloVe} & 0.63 & 0.47 & 0.63 & *** &  & 0.55 & 0.39 & 0.57 &  &  & 0.85 & 0.73 & 0.86 &  &  & 0.27 & 0.13 & 0.29 &  \\
\textsc{Dep-Embs} & 0.62 & 0.42 & 0.60 & *** &  & 0.54 & 0.36 & 0.54 & *** &  & 0.84 & 0.67 & 0.84 & * &  & 0.26 & 0.12 & 0.28 &  \\
\textsc{ELMo} & 0.67 & 0.55 & 0.67 & ** &  & 0.58 & 0.44 & 0.58 & *** &  & 0.84 & 0.71 & 0.85 & ** &  & 0.31 & 0.17 & 0.34 &  \\
\textsc{BERT} & \textbf{0.74} & \textbf{0.67} & \textbf{0.74} & $\leftarrow$ &  & \textbf{0.64} & \textbf{0.59} & \textbf{0.67} & $\leftarrow$ &  & \textbf{0.87} & \textbf{0.77} & \textbf{0.88} & $\leftarrow$ &  & \textbf{0.36} & 0.25 & \textbf{0.37} & $\leftarrow$ \\ \midrule
\textsc{Human} & 0.78 & 0.80 & 0.67 &  &  & 0.70 & 0.69 & 0.61 &  &  & 0.83 & 0.93 & 0.80 &  &  & 0.65 & 0.67 & 0.40 &  \\ \bottomrule
\end{tabular}
}
\caption{Macro F1 scores per category (object, property, affordance) and micro F1 score (\textit{$\mu$F1}) on both the abstract and situated test sets. Highest model values are bolded. Statistical significance (\textit{sig}) is calculated with McNemar's test, comparing the best-scoring model (by \textit{$\mu$F1}, denoted $\leftarrow$) with each other model. Stratified p-values are shown, with * for $p < 0.05$, ** for $p < 0.01$, and *** for $p < 0.001$. Human performance is estimated by 50 expert-annotated random samples from the test set (no McNemar's test).}
\label{table:results}
\end{table*}

\subsection{Tasks}

As shown at the bottom of Figure~\ref{fig:headlne}, our problem space naturally defines three edges in a graph. A property prediction task may attempt to produce the human-labeled set of properties given a new object (O$\rightarrow$P) \cite{lucy2017distributional}.
Predicting affordances can be done similarly: given a new object, can its top affordances be distinguished from others (O$\rightarrow$A)?
And finally, the troublesome but fertile edge between properties and affordances: can a model predict the set of properties compatible with an affordance (A$\rightarrow$P)?

We frame each scenario as a series of joint reasoning tasks.
Given two instances (e.g., an object and a property), a model must make a binary decision as to whether they are compatible.
For example, predicting which of $k$ properties $\{p_1,\ldots,p_k\}$ are compatible with an object $o$ will be set up as $k$ compatibility tasks $(o, p_i) \rightarrow \{0,1\}$.\footnote{We experimented with other task setups found in previous work, such as using an object to predict a $k$-length vector of properties: $(o) \rightarrow \{0,1\}_k$. However, we found models performed better on all metrics by instead framing the task as a series of compatibility decisions. We suspect the reason is that this setup allows models to take advantage of input representations of both words rather than just one.}
We denote the tasks as object-property (O$\longleftrightarrow$P), object-affordance (O$\longleftrightarrow$A), and affordance-property (A$\longleftrightarrow$P).

\subsection{Data}

To fuel experiments in these three tasks, we introduce two new datasets.
The first we call the \textit{abstract} dataset, which is a set of judgements elicited from only the name of the object (e.g., \textit{wheelbarrow}) and property (e.g., \textit{is an animal}).
The second is the \textit{situated} dataset, where properties and affordances are annotated on objects in the context of real-world pictures.\footnote{Annotations for both datasets are performed by workers on Amazon Mechanical Turk.}

\paragraph{Abstract Dataset}
Several lists of properties \cite{mcrae2005semantic}, categorization schemes \cite{devereux2014centre}, and quantification layers \cite{herbelot2015concepts} have been proposed. We take the set of objects and properties from \citeA{mcrae2005semantic} and perform filtering and preprocessing similar to \citeA{lucy2017distributional}.  We also include the set of objects from the MS COCO dataset \cite{lin2014microsoft}, collapse similar objects (e.g., many bird species) and add seven new properties (such as \textit{man-made} and \textit{squishy}). We end up with a set of 514 objects and 50 properties. We re-annotate all 25,700 object-property pairs to eliminate false negatives from the original McRae data collection process and provide labels for new entries. We annotate each pair three times for a total of 77,100 annotations, and keep only labels with $\geq 2/3$ agreement.

\paragraph{Situated Dataset}
We also annotate instances of objects situated in photographs.
Images have the great advantage of resolving visual ambiguities of appearance, shape, and form.
For example, a \textit{bottle} has different properties if it is a glass beverage container or plastic shampoo tube.
Only a few non-visual properties (e.g., \textit{smelliness}) must then be inferred from the environment.

To build the an experimental situated testbed, we sample images from the MS COCO dataset \cite{lin2014microsoft}. We constrain each image to have between three and seven objects to avoid scenes that are too sparse (often portraits) or dense (cluttered collections). We also ensure that we have at least five samples of each of the 80 unique object categories in the dataset. We end up with 1,024 objects across 220 images. We then annotate all 50 properties (introduced in the abstract dataset) for each object, annotating each three times for a total of 153,600 labels. We filter using the same scheme ($\geq 2/3$ agreement).

In addition to the properties, we also collect annotations of the affordances for all objects in the situated dataset.
We allow annotators to choose from the 504 verbs from the imSitu dataset \cite{yatskar2016situation}.
We provide common variants of each verb that include particles, allowing annotations such as \textit{pick up} and \textit{throw out}.
Annotators select the top three to five affordances that come to mind when they see the selected object in the context of its photograph.
We again perform this annotation three times for each object, and aggregate the verbs chosen to pick the top three most common affordances for each object.
We end up with a set of sparsely labeled affordances for each situated object.
We perform balanced negative sampling by selecting $k=3$ affordances for each datum and setting their labels to zero.

Detailed statistics and examples for both datasets are shown in Table~\ref{table:dataset-stats}. Full lists of the objects, properties, and affordances, as well as the annotation interfaces, are provided in the Appendix.

\subsection{Models}

\paragraph{Word embeddings}
We consider four representations of the words involved in the tasks. Two of the representations are word embeddings.
These map single words to vectors in $\mathbb{R}^d$.
We use GloVe embeddings \cite{pennington2014glove} as they have proven effective at object-property tasks in the past \cite{lucy2017distributional}.
We also use Dependency Based Word Embeddings \cite{levy2014dependency}, as they may more directly capture the relations between objects and their affordances.
In both cases, $d=300$, and we use the GloVe embedding variant with the largest amount of pretraining (840 billion words).

\paragraph{Contextualized representations}
The other two representations are ELMo \cite{peters2018deep} and BERT \cite{devlin2018bert}, which are contextualized.
These require full sentences (as opposed to single words) to compute a vector, but in turn produce results more specific to words' linguistic surroundings.
For example, ELMo and BERT produce different representations for \textit{book} in \textit{``I read the book''} versus \textit{``Please book the flight,''} while word embeddings have only a single representation.

To account for this, we generate sentences using the relevant objects, properties, and affordances for the task at hand.
For example, to judge \textit{accordion} and \textit{squishy}, we would generate \textit{``An accordion is squishy.''}

For ELMo, we then take the final layer representations for the two compared words, each of which is a $d=1024$ length vector.
For BERT, we take the overall sentence representation as the final layer's hidden state of the \texttt{[CLS]} (sentence summary) symbol, which produces a single $d=1024$ vector.

\paragraph{Finetuning}
Given the word representations above, we finetune each of the models by adding trainable multilayer perceptron (MLP) after the input representations.
This allows models to learn interrelations between the two categories at hand, essentially calibrating the unsupervised representations into a compatibility function.
We use a single hidden layer in the MLP, and train using mean squared error loss with L2 regularization. For BERT, we find the standard procedure of finetuning the \textit{entire} model vital for good performance.

To summarize, for two words $(w_i, w_j)$ which can be written together in a sentence $s = w_1 ... w_n$, we have for a model $m$,

$$
r(w_i,w_j) =
\left\{
	\begin{array}{ll}
		\langle m(w_i), m(w_j) \rangle  & \mbox{if } m \in \textsc{\{Gl., D.E.\}} \\
		m_{\{i,j\}}^{-1}(s) & \mbox{if } m = \textsc{ELMo} \\
		m^{-1}_{\texttt{[CLS]}}(s) & \mbox{if } m = \textsc{BERT}
	\end{array}
\right.
$$

$$
    \hat{y}_{w_i, w_j} \propto \sigma(\mathbf{W_2} \times a(\mathbf{W_1} \times r(w_i,w_j) + \mathbf{b_1}) + \mathbf{b_2}) \\
$$

$$
\mathcal{L}(w_i, w_j, y, \mathbf{\theta}, \lambda) = (y - \hat{y}_{w_i, w_j})^2 + \lambda \| \mathbf{\theta} \|^2_2
$$

\noindent where $m(\cdot)^\ell_i$ is an embedding of the $i$th token in the layer $\ell$, $a$ is a nonlinear activation function, $y \in \{0,1\}$ is the ground truth label, $\mathbf{\theta = \{W_1, W_2, b_1, b_2\}}$ are trainable parameters, and $\lambda$ is the regularization strength.\footnote{For BERT, we also follow standard practice and append a single trainable layer (logistic regression) instead of a two-layer MLP; i.e., $\mathbf{W_1} = \mathbf{I}, \mathbf{b_1} = \mathbf{0}, a(x) = x$.}

We optimize models using gradient descent (or Adam \cite{kingma2014adam} for BERT), and tune all hyperparameters using $k$-fold cross validation with $k=5$.

\begin{figure*}[t!]

\begin{center}
\includegraphics[width=0.99\linewidth]{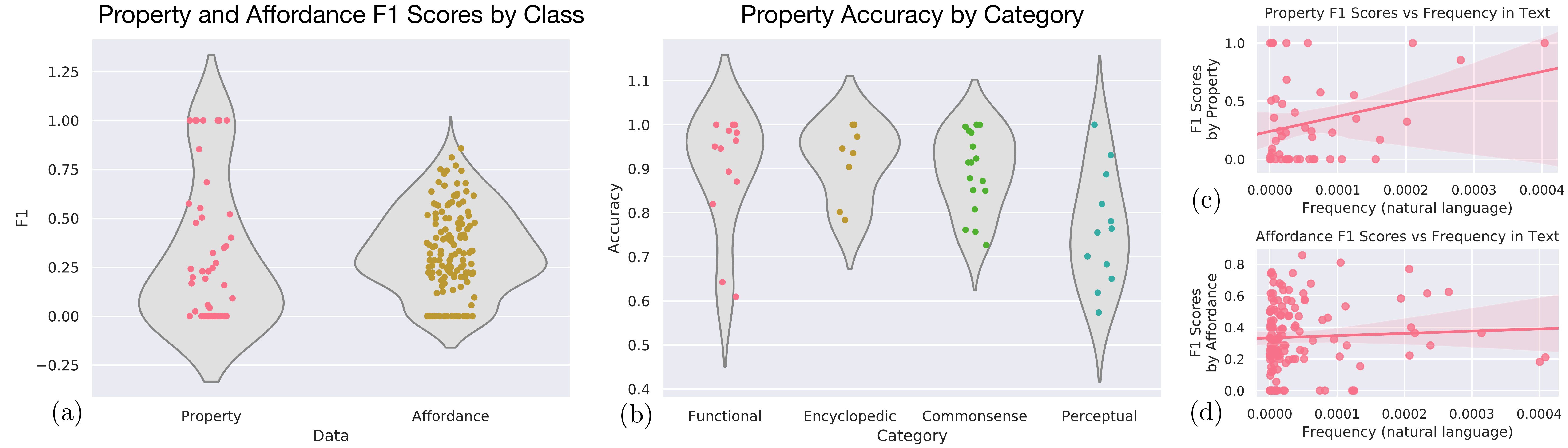}
\end{center}

\caption{Detailed results of top performing model (BERT) on the affordance-property compatibility task (A$\longleftrightarrow$P) in the situated dataset. (a) F1 scores are plotted per property (left) and affordance (right). (b) Properties are divided into four categories and plotted by accuracy. (c), (d) Both property and affordance F1 plotted against word frequency in natural language text.}
\label{fig:results-graphs}

\end{figure*}

\paragraph{Baselines} We compare performance for these models against two simple approaches. The \textit{random} baseline simply flips a coin for each compatibility decision. The \textit{majority} baseline uses the per-class majority label for the training set, aggregating by property for the $O \longleftrightarrow P$ and $A \longleftrightarrow P$ tasks, and by affordance for the $O \longleftrightarrow A$ task.

\paragraph{Human performance} Finally, we estimate human performance on this task. We sample 50 samples at random from the test set for each task, and have an expert annotate them. For fairness to the models, we do not show the expert the photographs or exact instance from which the situated examples are drawn.

\subsection{Results}

A summary of all model performances is shown in Table~\ref{table:results}.
Consistent with prior work that has studied object and property compatibility \cite{lucy2017distributional}, we find good but not perfect performance (close to 0.70 F1 scores) on the abstract dataset (task $O \longleftrightarrow P$).
Models fare slightly worse on the situated $O \longleftrightarrow P$ task, with the best performance below 0.60 F1.
This effect is consistent in the human expert scores as well.
Though this dataset is larger, the introduction of context allows for greater variance in the properties of an object.

The object-affordance compatibility task ($O\longleftrightarrow A$) yields significantly higher numbers.
Not only is this task statistically easier (as demonstrated by the strong majority baseline), but this edge is the only one directly observed in language. 
All models pretrained on text have been exposed to many instances of likely verbs for each object considered.
In fact, all pretrained models perform in the same range as human ability, and there is no statistically significant difference between the models for this task.

However, all models struggle with the affordance-property task ($A \longleftrightarrow P$).
The highest F1 scores are in the 0.30s, with the random baseline achieving the highest macro F1 score by property.
While this task is also the most difficult for humans, their macro F1 scores for both affordances and properties are around double those of the best performing models.
We posit that the inference between affordances and properties requires multi-hop reasoning that is simply not present in the pretraining of large text-based models.
We provide further analysis in the following section.
\section{Analysis}

Models achieve reasonable performance predicting the compatibility of both properties and affordances with objects.
However, the task requiring inference between affordances and properties (A$\longleftrightarrow$P) confounds even the strongest models.

We explore this result through a detailed analysis of the top performing model.
Figure~\ref{fig:results-graphs} presents a breakdown of BERT's results on the affordances-property compatibility task (A$\longleftrightarrow$P) on the situated dataset.
From the leftmost graph (a), we observe that a per-property analysis shows a largely bimodal split between properties that are fully predicted (1.0 F1), and went completely unmodeled  (0.0 F1).
Affordances, on the other hand, lie more evenly across the F1 range.
Because the task involved the compatibility between properties and affordances, mass for correct predictions must be shared between the two data groups.
That so few properties achieved a high F1 score suggests that many affordances rely on only a few properties for accurate prediction.

We perform further analysis to investigate which kinds of properties yielded better affordance-property modeling.
We categorize each property into four coarse classes: functional (e.g., \textit{is used for cooking}), encyclopedic (e.g., \textit{is an animal}), commonsense (e.g., \textit{comes in pairs}), and perceptual (e.g., \textit{is smooth}).
Figure ~\ref{fig:results-graphs} (b) shows a breakdown of property performance grouped by these four categories.
(Here, we plot accuracy instead of the sharper F1 metric to better illustrate the spread of performance.)
Functional properties exhibit the highest performance.
This makes intuitive since, because functional capabilities are directly tied to an object's affordances.
In contrast, perceptual properties exhibit generally lower and inconsistent performance than other categories.
We suspect that perceptual observations observed in text are not expressed with affordances, making this connection difficult for models.
Largely perceptual features can be written about with simple verbs (\textit{hear, see, feel}), giving them less implicit evidence than more nuanced properties.
Finally, encyclopedic and commonsense properties fall somewhere in the middle.
These properties, which involve an object's general characteristics (like \textit{requires gasoline, lives in water,} or \textit{has a peel}), correlate with a variety of verbs.
But they may only be directly expressed at a distance from a verb, making the inference between them still challenging.

Our final analyses in Figure~\ref{fig:results-graphs} (c) and (d) investigate whether there is a link between the predictive power of the model and how often a word is used in text.
We compute the frequencies of all affordances and properties occurring in natural language using the Google Web 1T corpus, an n-gram corpus computed from approximately one trillion words \cite{brants2006web}.
Figure~\ref{fig:results-graphs} (c) plots the F1 score of properties against how frequently they appear in natural language; \ref{fig:results-graphs} (d) plots the same for affordances.
We include a best-fit line along with confidence intervals shown as one standard deviation of the data.
We do not observe a statistical correlation between how much affordances and properties are written about, and how well neural models are able to connect their effects; a single confidence interval spans both positive and negative slopes.
This lack of clear correlation is surprising, because large state-of-the-art neural textual models generally improve with repeated exposure to instances of words.
Except for the three most common words measured by property F1 score, the rest of the data shows a strikingly uniform distribution of F1 scores for any choice of frequency in natural language.
This suggests that current neural models are fundamentally limited in their capacity for physical reasoning, and that only new designs---not more data---can allow them to acquire this skill.

\section{Discussion}

Despite being able to associate a considerable range of information with the names of objects, neural models are not able to capture the more subtle interplay between affordances and properties. In some sense, this result is unsurprising. Collecting information around an object can be informed largely by the co-occurrence of words around that object's various mentions. Affordances that imply properties (and the reverse) are rarely mentioned together; their mutual connotation naturally renders joint expression redundant. Hence, priorless models that learn from statistical associations falter. Given the depth of the networks used in models such as ELMo and BERT, complex inter-parameter structure arises, but the latent semantic patterns that describe physical commonsense are much weaker than more superficial patterns that arise due to grammar or domain.

This evidence evokes theories of embodied cognition \cite{gover1996embodied, wilson2002six}, which suggest that the nature of human cognition depends strongly on the stimuli granted by physical experience. If this is so, then how is information encoded in our physical experience such that we can make predictions? If we assume a form of mental simulation, then what are the mental limits on its reliability? From an artificial intelligence perspective, the more interesting proof is in the principles of creating such a mental simulator. If we are to simulate human capacity for thought, how actually must we simulate elements of the physical world?

With the rise of physics engines, our ability to model physical inferences grows \cite{wu2015galileo}. However, while this may make us better at anticipating human predictions about physical situations through perceptual stimuli \cite{gerstenbergfaulty}, there is still a long way to go before we understand the inferences that are being made through more symbolic stimuli, such as language. Exploring the mechanisms underlying this communication using an implicit shared world model will require us to either develop access to such a world model, or expose algorithms to predictions of that world model by directly querying humans. Bridging the inductive biases learned from simulation \cite{battaglia2013simulation} and those discovered by scientists \cite{lake2019human} to make inferences implicit in text will lead to a more cohesive model of commonsense physics. We expect such a model to bear fruit in studies of communication rich with physical implications.
\section{Acknowledgments}
This work was supported by NSF grants (IIS-1524371, 1637479, 1703166), NSF Fellowship, the DARPA CwC program through ARO (W911NF-15-1-0543), and gifts by Google and Facebook. The views and conclusions contained herein are those of the authors and should not be interpreted as representing endorsements of the funding agencies. 

\bibliographystyle{apacite}

\setlength{\bibleftmargin}{.125in}
\setlength{\bibindent}{-\bibleftmargin}

\bibliography{main}

\newpage
\section{Appendix}

\subsection{Objects}

We provide below a full list of the objects considered in both of our datasets. We note the split that it belongs to in each dataset (train or test), and the origin of the object (MR = \citeA{mcrae2005semantic}; C = MS COCO \cite{lin2014microsoft}).

In general, this list is the union of objects found in McRae \citeyear{mcrae2005semantic} and MS COCO. For cases where we do not use the object in either dataset (i.e., a ``-'' in both columns), we mark the row in italics and provide a note for why it was dropped. The most common reasons for dropping an object are: \textbf{polysemy}, such as in \emph{bat (animal \emph{vs} baseball)}; \textbf{hypernomy} (e.g., \textit{sparrow $\rightarrow$ bird}) to collapse similar objects, such as the eighteen species of birds found in McRae \citeyear{mcrae2005semantic}; and \textbf{dialect} (e.g., \textit{trousers $\rightarrow$ pants}) to make annotations easier for readers of American English. (For the several species of fish in McRae \citeyear{mcrae2005semantic}, we left two fish of clearly distinct sizes, \textit{goldfish} and \textit{trout}, rather than including a generic \textit{``fish''} object.)

\vspace{2em}
\begin{footnotesize}
\tablehead{\toprule \textbf{Object} & \textbf{Abstract} & \textbf{Situated} & \textbf{Origin} & \textbf{Note}\\\toprule}
\begin{supertabular}{@{}|lllll|@{}}
accordion & train & - & MR &  \\
airplane & train & test & MR, C &  \\
alligator & train & - & MR &  \\
ambulance & train & - & MR &  \\
anchor & train & - & MR &  \\
ant & train & - & MR &  \\
apartment & train & - & MR &  \\
apple & train & train & MR, C &  \\
apron & train & - & MR &  \\
armour & test & - & MR &  \\
ashtray & train & - & MR &  \\
asparagus & train & - & MR &  \\
avocado & train & - & MR &  \\
axe & test & - & MR &  \\
backpack & test & train & C &  \\
bag & train & - & MR &  \\
bagpipe & train & - & MR &  \\
ball & train & - & MR &  \\
balloon & train & - & MR &  \\
banana & train & train & MR, C &  \\
banjo & test & - & MR &  \\
banner & train & - & MR &  \\
barn & train & - & MR &  \\
barrel & train & - & MR &  \\
baseball bat & train & train & C &  \\
baseball glove & test & train & C &  \\
basement & train & - & MR &  \\
basket & train & - & MR &  \\
\textit{bat (animal)} & \textit{-} & \textit{-} & \textit{MR} & \textit{(polysemy)} \\
\textit{bat (baseball)} & \textit{-} & \textit{-} & \textit{MR} & \textit{(polysemy)} \\
bathtub & test & - & MR &  \\
baton & train & - & MR &  \\
bayonet & train & - & MR &  \\
bazooka & train & - & MR &  \\
beans & train & - & MR &  \\
bear & train & train & MR, C &  \\
beaver & test & - & MR &  \\
bed & train & train & MR, C &  \\
bedroom & train & - & MR &  \\
beehive & train & - & MR &  \\
beetle & train & - & MR &  \\
beets & train & - & MR &  \\
belt & train & - & MR &  \\
bench & train & train & MR, C &  \\
bicycle & test & train & C &  \\
bike & train & - & MR &  \\
\textit{bin (waste)} & \textit{-} & \textit{-} & \textit{MR} & \textit{(polysemy)} \\
\textit{birch} & \textit{-} & \textit{-} & \textit{MR} & \textit{$\rightarrow$ tree} \\
bird & train & train & C &  \\
biscuit & train & - & MR &  \\
bison & train & - & MR &  \\
\textit{blackbird} & \textit{-} & \textit{-} & \textit{MR} & \textit{$\rightarrow$ bird} \\
blender & train & - & MR &  \\
blouse & train & - & MR &  \\
blueberry & train & - & MR &  \\
\textit{bluejay} & \textit{-} & \textit{-} & \textit{MR} & \textit{$\rightarrow$ bird} \\
\textit{board (black)} & \textit{-} & \textit{-} & \textit{MR} & \textit{(polysemy)} \\
\textit{board (wood)} & \textit{-} & \textit{-} & \textit{MR} & \textit{(polysemy)} \\
boat & train & train & MR, C &  \\
bolts & train & - & MR &  \\
bomb & train & - & MR &  \\
book & test & train & MR, C &  \\
bookcase & train & - & MR &  \\
boots & train & - & MR &  \\
bottle & train & test & MR, C &  \\
bouquet & train & - & MR &  \\
\textit{bow (ribbon)} & \textit{-} & \textit{-} & \textit{MR} & \textit{(polysemy)} \\
\textit{bow (weapon)} & \textit{-} & \textit{-} & \textit{MR} & \textit{(polysemy)} \\
bowl & train & test & MR, C &  \\
box & train & - & MR &  \\
bra & train & - & MR &  \\
bracelet & train & - & MR &  \\
bread & train & - & MR &  \\
brick & train & - & MR &  \\
bridge & train & - & MR &  \\
broccoli & train & test & MR, C &  \\
broom & train & - & MR &  \\
brush & test & - & MR &  \\
bucket & test & - & MR &  \\
buckle & test & - & MR &  \\
\textit{budgie} & \textit{-} & \textit{-} & \textit{MR} & \textit{(obscure)} \\
buffalo & train & - & MR &  \\
buggy & train & - & MR &  \\
building & test & - & MR &  \\
bull & train & - & MR &  \\
bullet & train & - & MR &  \\
bungalow & train & - & MR &  \\
bureau & train & - & MR &  \\
bus & train & train & MR, C &  \\
butterfly & train & - & MR &  \\
\textit{buzzard} & \textit{-} & \textit{-} & \textit{MR} & \textit{$\rightarrow$ bird} \\
cabbage & test & - & MR &  \\
cabin & train & - & MR &  \\
cabinet & train & - & MR &  \\
cage & train & - & MR &  \\
cake & train & train & MR, C &  \\
calf & train & - & MR &  \\
camel & train & - & MR &  \\
camisole & train & - & MR &  \\
\textit{canary} & \textit{-} & \textit{-} & \textit{MR} & \textit{$\rightarrow$ bird} \\
candle & train & - & MR &  \\
cannon & test & - & MR &  \\
canoe & train & - & MR &  \\
cantaloupe & train & - & MR &  \\
\textit{cap (bottle)} & \textit{-} & \textit{-} & \textit{MR} & \textit{(polysemy)} \\
\textit{cap (hat)} & \textit{-} & \textit{-} & \textit{MR} & \textit{(polysemy)} \\
cape & train & - & MR &  \\
car & train & train & MR, C &  \\
\textit{card (greeting)} & \textit{-} & \textit{-} & \textit{MR} & \textit{(polysemy)} \\
caribou & train & - & MR &  \\
carpet & train & - & MR &  \\
carrot & train & train & MR, C &  \\
cart & test & - & MR &  \\
cat & train & train & MR, C &  \\
catapult & train & - & MR &  \\
caterpillar & train & - & MR &  \\
\textit{catfish} & \textit{-} & \textit{-} & \textit{MR} & \textit{(obscure)} \\
cathedral & train & - & MR &  \\
cauliflower & test & - & MR &  \\
\textit{cedar} & \textit{-} & \textit{-} & \textit{MR} & \textit{$\rightarrow$ tree} \\
celery & train & - & MR &  \\
cell phone & train & train & C &  \\
cellar & train & - & MR &  \\
cello & train & - & MR &  \\
certificate & train & - & MR &  \\
chain & train & - & MR &  \\
chair & test & test & MR, C &  \\
chandelier & train & - & MR &  \\
chapel & train & - & MR &  \\
cheese & test & - & MR &  \\
cheetah & train & - & MR &  \\
cherry & train & - & MR &  \\
\textit{chickadee} & \textit{-} & \textit{-} & \textit{MR} & \textit{$\rightarrow$ bird} \\
chicken & test & - & MR &  \\
chimp & train & - & MR &  \\
chipmunk & train & - & MR &  \\
chisel & train & - & MR &  \\
church & train & - & MR &  \\
cigar & train & - & MR &  \\
cigarette & train & - & MR &  \\
clam & train & - & MR &  \\
clamp & train & - & MR &  \\
clarinet & train & - & MR &  \\
cloak & train & - & MR &  \\
clock & train & train & MR, C &  \\
closet & train & - & MR &  \\
coat & train & - & MR &  \\
cockroach & train & - & MR &  \\
coconut & train & - & MR &  \\
\textit{cod} & \textit{-} & \textit{-} & \textit{MR} & \textit{(obscure)} \\
coin & train & - & MR &  \\
colander & train & - & MR &  \\
comb & test & - & MR &  \\
cork & train & - & MR &  \\
corkscrew & train & - & MR &  \\
corn & test & - & MR &  \\
cottage & train & - & MR &  \\
couch & train & train & MR, C &  \\
cougar & train & - & MR &  \\
cow & test & train & MR, C &  \\
coyote & train & - & MR &  \\
crab & test & - & MR &  \\
cranberry & train & - & MR &  \\
\textit{crane (machine)} & \textit{-} & \textit{-} & \textit{MR} & \textit{(polysemy)} \\
crayon & train & - & MR &  \\
crocodile & train & - & MR &  \\
crossbow & train & - & MR &  \\
crow & train & - & MR &  \\
crowbar & test & - & MR &  \\
crown & train & - & MR &  \\
cucumber & test & - & MR &  \\
cup & train & train & MR, C &  \\
cupboard & test & - & MR &  \\
curtains & train & - & MR &  \\
cushion & test & - & MR &  \\
dagger & train & - & MR &  \\
dandelion & train & - & MR &  \\
deer & test & - & MR &  \\
desk & train & - & MR &  \\
dining table & train & test & C &  \\
dish & train & - & MR &  \\
dishwasher & train & - & MR &  \\
dog & train & train & MR, C &  \\
doll & test & - & MR &  \\
dolphin & train & - & MR &  \\
donkey & train & - & MR &  \\
donut & train & train & C &  \\
door & train & - & MR &  \\
doorknob & train & - & MR &  \\
dove & train & - & MR &  \\
drain & train & - & MR &  \\
drapes & train & - & MR &  \\
dress & test & - & MR &  \\
dresser & train & - & MR &  \\
drill & train & - & MR &  \\
drum & train & - & MR &  \\
duck & train & - & MR &  \\
\textit{dunebuggy} & \textit{-} & \textit{-} & \textit{MR} & \textit{(no word emb.)} \\
eagle & test & - & MR &  \\
earmuffs & train & - & MR &  \\
eel & train & - & MR &  \\
eggplant & train & - & MR &  \\
elephant & test & test & MR, C &  \\
elevator & train & - & MR &  \\
elk & test & - & MR &  \\
emerald & train & - & MR &  \\
\textit{emu} & \textit{-} & \textit{-} & \textit{MR} & \textit{(obscure)} \\
envelope & train & - & MR &  \\
escalator & train & - & MR &  \\
falcon & train & - & MR &  \\
\textit{fan (appliance)} & \textit{-} & \textit{-} & \textit{MR} & \textit{(polysemy)} \\
faucet & train & - & MR &  \\
fawn & train & - & MR &  \\
fence & train & - & MR &  \\
\textit{finch} & \textit{-} & \textit{-} & \textit{MR} & \textit{$\rightarrow$ bird} \\
fire hydrant & train & train & C &  \\
flamingo & test & - & MR &  \\
flea & train & - & MR &  \\
flute & train & - & MR &  \\
football & train & - & MR &  \\
fork & train & train & MR, C &  \\
fox & train & - & MR &  \\
freezer & train & - & MR &  \\
fridge & train & - & MR &  \\
frisbee & train & train & C &  \\
frog & train & - & MR &  \\
garage & train & - & MR &  \\
garlic & train & - & MR &  \\
gate & test & - & MR &  \\
giraffe & train & train & MR, C &  \\
gloves & train & - & MR &  \\
goat & train & - & MR &  \\
goldfish & train & - & MR &  \\
goose & train & - & MR &  \\
gopher & train & - & MR &  \\
gorilla & train & - & MR &  \\
gown & test & - & MR &  \\
grape & train & - & MR &  \\
grapefruit & train & - & MR &  \\
grasshopper & train & - & MR &  \\
grater & train & - & MR &  \\
grenade & train & - & MR &  \\
groundhog & train & - & MR &  \\
guitar & train & - & MR &  \\
gun & train & - & MR &  \\
\textit{guppy} & \textit{-} & \textit{-} & \textit{MR} & \textit{(obscure)} \\
hair drier & train & test & C &  \\
hammer & train & - & MR &  \\
hamster & train & - & MR &  \\
handbag & train & train & C &  \\
hare & train & - & MR &  \\
harmonica & train & - & MR &  \\
harp & train & - & MR &  \\
harpoon & train & - & MR &  \\
harpsichord & train & - & MR &  \\
hatchet & train & - & MR &  \\
hawk & test & - & MR &  \\
helicopter & test & - & MR &  \\
helmet & train & - & MR &  \\
hoe & train & - & MR &  \\
honeydew & train & - & MR &  \\
hook & train & - & MR &  \\
hornet & train & - & MR &  \\
horse & train & test & MR, C &  \\
hose & train & - & MR &  \\
\textit{hose (leggings)} & \textit{-} & \textit{-} & \textit{MR} & \textit{(polysemy)} \\
hot dog & train & test & C &  \\
house & test & - & MR &  \\
housefly & test & - & MR &  \\
hut & test & - & MR &  \\
hyena & train & - & MR &  \\
iguana & test & - & MR &  \\
inn & test & - & MR &  \\
jacket & test & - & MR &  \\
jar & train & - & MR &  \\
jeans & train & - & MR &  \\
jeep & train & - & MR &  \\
jet & test & - & MR &  \\
kettle & train & - & MR &  \\
key & train & - & MR &  \\
keyboard & train & test & C &  \\
\textit{keyboard (musical)} & \textit{-} & \textit{-} & \textit{MR} & \textit{(polysemy)} \\
kite & train & train & MR, C &  \\
knife & train & train & MR, C &  \\
ladle & train & - & MR &  \\
lamb & train & - & MR &  \\
lamp & train & - & MR &  \\
lantern & test & - & MR &  \\
laptop & test & train & C &  \\
lemon & train & - & MR &  \\
leopard & train & - & MR &  \\
leotards & train & - & MR &  \\
lettuce & train & - & MR &  \\
level & train & - & MR &  \\
lime & test & - & MR &  \\
limousine & train & - & MR &  \\
lion & train & - & MR &  \\
lobster & test & - & MR &  \\
machete & train & - & MR &  \\
\textit{mackerel} & \textit{-} & \textit{-} & \textit{MR} & \textit{(obscure)} \\
magazine & test & - & MR &  \\
mandarin & test & - & MR &  \\
marble & train & - & MR &  \\
mat & train & - & MR &  \\
medal & train & - & MR &  \\
menu & train & - & MR &  \\
microscope & train & - & MR &  \\
microwave & train & train & MR, C &  \\
\textit{mink} & \textit{-} & \textit{-} & \textit{MR} & \textit{(polysemy)} \\
\textit{mink (coat)} & \textit{-} & \textit{-} & \textit{MR} & \textit{(polysemy)} \\
\textit{minnow} & \textit{-} & \textit{-} & \textit{MR} & \textit{(obscure)} \\
mirror & train & - & MR &  \\
missile & train & - & MR &  \\
mittens & test & - & MR &  \\
mixer & train & - & MR &  \\
\textit{mole (animal)} & \textit{-} & \textit{-} & \textit{MR} & \textit{(polysemy)} \\
moose & train & - & MR &  \\
moth & train & - & MR &  \\
motorcycle & test & train & MR, C &  \\
mouse & train & train & MR, C &  \\
\textit{mouse (computer)} & \textit{-} & \textit{-} & \textit{MR} & \textit{(polysemy)} \\
mug & train & - & MR &  \\
mushroom & train & - & MR &  \\
muzzle & train & - & MR &  \\
napkin & train & - & MR &  \\
necklace & train & - & MR &  \\
nectarine & train & - & MR &  \\
nightgown & train & - & MR &  \\
\textit{nightingale} & \textit{-} & \textit{-} & \textit{MR} & \textit{$\rightarrow$ bird} \\
nylons & test & - & MR &  \\
\textit{oak} & \textit{-} & \textit{-} & \textit{MR} & \textit{$\rightarrow$ tree} \\
octopus & test & - & MR &  \\
olive & train & - & MR &  \\
onions & test & - & MR &  \\
orange & train & train & MR, C &  \\
\textit{oriole} & \textit{-} & \textit{-} & \textit{MR} & \textit{$\rightarrow$ bird} \\
ostrich & test & - & MR &  \\
otter & train & - & MR &  \\
oven & test & train & MR, C &  \\
owl & train & - & MR &  \\
ox & test & - & MR &  \\
paintbrush & train & - & MR &  \\
pajamas & train & - & MR &  \\
pan & train & - & MR &  \\
panther & train & - & MR &  \\
pants & train & - & MR &  \\
\textit{parakeet} & \textit{-} & \textit{-} & \textit{MR} & \textit{$\rightarrow$ bird} \\
parka & train & - & MR &  \\
parking meter & train & train & C &  \\
parsley & train & - & MR &  \\
\textit{partridge} & \textit{-} & \textit{-} & \textit{MR} & \textit{$\rightarrow$ bird} \\
peach & test & - & MR &  \\
peacock & train & - & MR &  \\
pear & train & - & MR &  \\
pearl & test & - & MR &  \\
peas & train & - & MR &  \\
peg & test & - & MR &  \\
\textit{pelican} & \textit{-} & \textit{-} & \textit{MR} & \textit{$\rightarrow$ bird} \\
pen & test & - & MR &  \\
pencil & train & - & MR &  \\
penguin & train & - & MR &  \\
pepper & test & - & MR &  \\
\textit{perch} & \textit{-} & \textit{-} & \textit{MR} & \textit{(obscure)} \\
person & train & train & C &  \\
\textit{pheasant} & \textit{-} & \textit{-} & \textit{MR} & \textit{$\rightarrow$ bird} \\
piano & train & - & MR &  \\
pickle & train & - & MR &  \\
pie & test & - & MR &  \\
pier & train & - & MR &  \\
pig & train & - & MR &  \\
pigeon & train & - & MR &  \\
pillow & test & - & MR &  \\
pin & train & - & MR &  \\
\textit{pine} & \textit{-} & \textit{-} & \textit{MR} & \textit{$\rightarrow$ tree} \\
pineapple & train & - & MR &  \\
\textit{pipe} & \textit{-} & \textit{-} & \textit{MR} & \textit{(polysemy)} \\
\textit{pipe (smoking)} & \textit{-} & \textit{-} & \textit{MR} & \textit{(polysemy)} \\
pistol & test & - & MR &  \\
pizza & test & train & C &  \\
plate & test & - & MR &  \\
platypus & test & - & MR &  \\
pliers & train & - & MR &  \\
\textit{plug (electric)} & \textit{-} & \textit{-} & \textit{MR} & \textit{(polysemy)} \\
plum & train & - & MR &  \\
pony & train & - & MR &  \\
porcupine & train & - & MR &  \\
pot & train & - & MR &  \\
potato & train & - & MR &  \\
potted plant & train & train & C &  \\
projector & train & - & MR &  \\
prune & train & - & MR &  \\
pumpkin & train & - & MR &  \\
pyramid & test & - & MR &  \\
python & train & - & MR &  \\
rabbit & train & - & MR &  \\
raccoon & train & - & MR &  \\
racquet & train & - & MR &  \\
radio & train & - & MR &  \\
radish & train & - & MR &  \\
raft & train & - & MR &  \\
raisin & test & - & MR &  \\
rake & test & - & MR &  \\
raspberry & train & - & MR &  \\
rat & train & - & MR &  \\
rattle & train & - & MR &  \\
rattlesnake & train & - & MR &  \\
\textit{raven} & \textit{-} & \textit{-} & \textit{MR} & \textit{$\rightarrow$ bird} \\
razor & train & - & MR &  \\
refrigerator & train & train & C &  \\
remote & train & train & C &  \\
revolver & train & - & MR &  \\
rhubarb & train & - & MR &  \\
rice & train & - & MR &  \\
rifle & train & - & MR &  \\
\textit{ring (jewelry)} & \textit{-} & \textit{-} & \textit{MR} & \textit{(polysemy)} \\
robe & train & - & MR &  \\
\textit{robin} & \textit{-} & \textit{-} & \textit{MR} & \textit{$\rightarrow$ bird} \\
rock & test & - & MR &  \\
rocker & test & - & MR &  \\
rocket & train & - & MR &  \\
rooster & train & - & MR &  \\
rope & test & - & MR &  \\
ruler & train & - & MR &  \\
sack & train & - & MR &  \\
saddle & train & - & MR &  \\
sailboat & test & - & MR &  \\
salamander & train & - & MR &  \\
salmon & train & - & MR &  \\
sandals & train & - & MR &  \\
sandpaper & train & - & MR &  \\
sandwich & test & train & C &  \\
sardine & test & - & MR &  \\
\textit{saucer} & \textit{-} & \textit{-} & \textit{MR} & \textit{$\rightarrow$ plate} \\
saxophone & test & - & MR &  \\
scarf & train & - & MR &  \\
scissors & train & train & MR, C &  \\
scooter & train & - & MR &  \\
screwdriver & train & - & MR &  \\
screws & test & - & MR &  \\
seagull & train & - & MR &  \\
seal & train & - & MR &  \\
seaweed & test & - & MR &  \\
shack & train & - & MR &  \\
shawl & train & - & MR &  \\
shed & train & - & MR &  \\
sheep & train & test & MR, C &  \\
shell & test & - & MR &  \\
shelves & train & - & MR &  \\
shield & train & - & MR &  \\
ship & train & - & MR &  \\
shirt & train & - & MR &  \\
shoes & train & - & MR &  \\
shotgun & train & - & MR &  \\
shovel & train & - & MR &  \\
shrimp & test & - & MR &  \\
sink & test & train & MR, C &  \\
skateboard & test & test & MR, C &  \\
skillet & train & - & MR &  \\
skirt & train & - & MR &  \\
skis & train & train & MR, C &  \\
skunk & train & - & MR &  \\
skyscraper & train & - & MR &  \\
sled & train & - & MR &  \\
sledgehammer & train & - & MR &  \\
sleigh & train & - & MR &  \\
slingshot & train & - & MR &  \\
slippers & train & - & MR &  \\
snail & train & - & MR &  \\
snowboard & test & train & C &  \\
socks & train & - & MR &  \\
sofa & test & - & MR &  \\
spade & train & - & MR &  \\
\textit{sparrow} & \textit{-} & \textit{-} & \textit{MR} & \textit{$\rightarrow$ bird} \\
spatula & train & - & MR &  \\
spear & train & - & MR &  \\
spider & train & - & MR &  \\
spinach & train & - & MR &  \\
spoon & test & train & MR, C &  \\
sports ball & train & train & C &  \\
squid & train & - & MR &  \\
squirrel & train & - & MR &  \\
\textit{starling} & \textit{-} & \textit{-} & \textit{MR} & \textit{$\rightarrow$ bird} \\
stereo & train & - & MR &  \\
stick & train & - & MR &  \\
stone & test & - & MR &  \\
\textit{stool (furniture)} & \textit{-} & \textit{-} & \textit{MR} & \textit{(polysemy)} \\
stop sign & train & train & C &  \\
\textit{stork} & \textit{-} & \textit{-} & \textit{MR} & \textit{$\rightarrow$ bird} \\
stove & train & - & MR &  \\
strainer & train & - & MR &  \\
strawberry & train & - & MR &  \\
submarine & test & - & MR &  \\
subway & train & - & MR &  \\
suitcase & train & train & C &  \\
surfboard & train & train & MR, C &  \\
swan & train & - & MR &  \\
sweater & train & - & MR &  \\
swimsuit & train & - & MR &  \\
sword & train & - & MR &  \\
table & train & - & MR &  \\
tack & test & - & MR &  \\
tangerine & train & - & MR &  \\
\textit{tank (army)} & \textit{-} & \textit{-} & \textit{MR} & \textit{(polysemy)} \\
\textit{tank (container)} & \textit{-} & \textit{-} & \textit{MR} & \textit{(polysemy)} \\
\textit{tap} & \textit{-} & \textit{-} & \textit{MR} & \textit{$\rightarrow$ faucet} \\
\textit{tape (scotch)} & \textit{-} & \textit{-} & \textit{MR} & \textit{(polysemy)} \\
taxi & train & - & MR &  \\
teddy bear & train & train & C &  \\
telephone & test & - & MR &  \\
tennis racket & train & test & C &  \\
tent & train & - & MR &  \\
thermometer & train & - & MR &  \\
thimble & train & - & MR &  \\
tie & train & train & MR, C &  \\
tiger & train & - & MR &  \\
toad & train & - & MR &  \\
toaster & train & train & MR, C &  \\
toilet & train & train & MR, C &  \\
tomahawk & train & - & MR &  \\
tomato & test & - & MR &  \\
tongs & train & - & MR &  \\
toothbrush & train & train & C &  \\
tortoise & train & - & MR &  \\
toy & test & - & MR &  \\
tractor & train & - & MR &  \\
traffic light & train & train & C &  \\
trailer & train & - & MR &  \\
train & test & train & MR, C &  \\
tray & train & - & MR &  \\
tree & train & - & new &  \\
tricycle & train & - & MR &  \\
tripod & train & - & MR &  \\
trolley & train & - & MR &  \\
trombone & train & - & MR &  \\
\textit{trousers} & \textit{-} & \textit{-} & \textit{MR} & \textit{$\rightarrow$ pants} \\
trout & train & - & MR &  \\
truck & train & test & MR, C &  \\
trumpet & train & - & MR &  \\
tuba & train & - & MR &  \\
tuna & train & - & MR &  \\
turkey & train & - & MR &  \\
turnip & train & - & MR &  \\
turtle & train & - & MR &  \\
tv & train & train & C &  \\
typewriter & train & - & MR &  \\
umbrella & train & train & MR, C &  \\
unicycle & train & - & MR &  \\
urn & train & - & MR &  \\
van & train & - & MR &  \\
vase & train & train & C &  \\
veil & test & - & MR &  \\
vest & test & - & MR &  \\
vine & train & - & MR &  \\
violin & train & - & MR &  \\
\textit{vulture} & \textit{-} & \textit{-} & \textit{MR} & \textit{$\rightarrow$ bird} \\
wagon & train & - & MR &  \\
wall & train & - & MR &  \\
walnut & train & - & MR &  \\
walrus & train & - & MR &  \\
wand & test & - & MR &  \\
wasp & train & - & MR &  \\
whale & test & - & MR &  \\
wheel & train & - & MR &  \\
wheelbarrow & train & - & MR &  \\
whip & train & - & MR &  \\
whistle & train & - & MR &  \\
\textit{willow} & \textit{-} & \textit{-} & \textit{MR} & \textit{$\rightarrow$ tree} \\
wine glass & train & train & C &  \\
woodpecker & train & - & MR &  \\
worm & test & - & MR &  \\
wrench & train & - & MR &  \\
yacht & train & - & MR &  \\
yam & train & - & MR &  \\
zebra & test & test & MR, C &  \\
zucchini & train & - & MR &  \\

\bottomrule
\end{supertabular} 
\end{footnotesize}

\vspace{4em}

\subsection{Properties}

We provide below the list of fifty properties used in both the abstract and situated datasets. We also list our categorization of the property into four areas: encyclopedic, perceptual, functional, or commonsense. Forty-three properties come from MR \cite{mcrae2005semantic}, and seven are new.

To build this list of properties, we started by the now-standard practice of filtering the properties used in \cite{mcrae2005semantic} to those that occur at least five times, which yields 266 properties. From this list, we selected properties that seemed applicable to a moderate range of objects. For example, we preferred \textit{produces noise} over the much more specific \textit{used by blowing air through}, and we added \textit{has words on it} over the near-unanimous \textit{has different colors}. We also avoided properties that were narrowly identifiable (e.g., \textit{has feet}) in favor of judgment-based properties (e.g., \textit{is slimy}). Finally, we added a few natural complements to the selected properties; e.g., from \textit{is usually cold} we included \textit{is usually hot}.

\vspace{2em}
\begin{footnotesize}
\tablehead{\toprule \textbf{Property} & \textbf{Categorization} & \textbf{Origin} \\ \toprule}
\begin{supertabular}{@{}|lll|@{}}
is an animal & encyclopedic & MR \\
is big & perceptual & MR \\
is breakable & commonsense & MR \\
is used by children & functional & MR \\
is used for cleaning & functional & MR \\
is usually cold & commonsense & MR \\
is used for cooking & functional & MR \\
is dangerous & commonsense & MR \\
is decorative & commonsense & MR \\
is used for eating & functional & MR \\
is edible & functional & MR \\
requires electricity & encyclopedic & MR \\
is expensive & commonsense & MR \\
is fast & commonsense & MR \\
is worn on feet & functional & MR \\
can fly & encyclopedic & MR \\
is fun & commonsense & MR \\
requires gasoline & encyclopedic & MR \\
is hand-held & commonsense & MR \\
is hard & perceptual & MR \\
is heavy & perceptual & MR \\
is used for holding things & functional & MR \\
is usually hot & commonsense & new \\
is used for killing & functional & MR \\
is light (in weight) & perceptual & new \\
is loud & perceptual & MR \\
is man-made & encyclopedic & new \\
is used for music & functional & MR \\
comes in pairs & commonsense & MR \\
has a peel & encyclopedic & MR \\
is sharp & commonsense & MR \\
has shelves & commonsense & MR \\
is shiny & perceptual & MR \\
is slimy & perceptual & MR \\
is smelly & perceptual & MR \\
is smooth & perceptual & MR \\
produces sound & perceptual & MR \\
is squishy & perceptual & new \\
is eaten in summer & commonsense & MR \\
can swim & encyclopedic & MR \\
is tall & perceptual & MR \\
is a tool & functional & MR \\
is a toy & functional & MR \\
is used for transportation & functional & MR \\
is unhealthy & commonsense & new \\
is found on walls & commonsense & MR \\
is worn for warmth & functional & MR \\
lives in water & encyclopedic & MR \\
is usually wet & commonsense & new \\
has words on it & commonsense & new \\
\bottomrule
\end{supertabular} 
\end{footnotesize}

\vspace{4em}

\subsection{Affordances}

We provide below the list of candidate verbs, along with the assistive particles and prepositions, we used for annotating affordances.

When annotating \textit{``What might you do to the X?''} for an object \textit{X}, it can be helpful or necessary to use a particle or preposition when writing an answer. Here are some examples:

\begin{align*}
    & \text{\textbf{Feed} the dog.} & \hspace{0.1em} & \textit{(verb only)} \\
    & \text{\textbf{Take out} the trash.} & \hspace{0.1em} & \textit{(verb + particle)} \\
    & \text{\textbf{Dive into} the water.} & \hspace{0.1em} & \textit{(verb + preposition)} \\
\end{align*}

We use two strategies to enable annotators to write grammatical constructions like the above. For particles, we provide common variants of each verb that use particles, such as \textit{buckle up} and \textit{buckle in} in addition to \textit{buckle}. For prepositions, we allow an additional choice of a preposition after the annotator has selected a verb (or verb + particle) from the list.

We discard both particles and prepositions when building our task data. We do this for two reasons. First, we wish to eliminate errors where an annotator mistakes the subtle distinction between a particle and a preposition. Second, we want to constrain the input and output space for the models, which would otherwise multiplicatively scale the verbs by the number of particles and prepositions.

To pick the verbs, we take the set of 504 verbs used in the imSitu dataset \cite{yatskar2016situation} and lemmatize them. For particles and prepositions, we run a dependency parser on a large corpus of sentences and aggregate statistics. We add all \textit{verb + particle} forms of all verbs that occur in at least 5\% of the usages of that verb. We then provide the twenty-six widely used prepositions as an additional selection. For brevity, we list here just the verbs in their lemmatized form, and provide the twenty-seven unique particles and prepositions.

\paragraph{Verbs}
\begin{footnotesize}
\textit{
adjust, admire, ail, aim, applaud, apply, apprehend, arch, arrange, arrest, ascend, ask,
assemble, attach, attack, autograph, bake, balloon, bandage, baptize, barbecue, bathe,
beg, bet, bike, bite, block, blossom, board, boat, bother, bounce, bow, braid, branch,
brawl, break, brew, browse, brush, bubble, buckle, build, bulldoze, burn, bury, butt,
butter, button, buy, call, calm, camouflage, camp, caress, carry, cart, carve, catch,
celebrate, chase, check, cheer, cheerlead, chew, chisel, chop, circle, clap, claw,
clean, clear, clench, climb, cling, clip, coach, collide, color, comb, communicate,
commute, compete, complain, confront, congregate, construct, cook, cough, count, cover,
craft, cram, crash, crawl, crest, crouch, crown, crush, cry, curl, curtsy, dance,
decompose, decorate, deflect, descend, destroy, detain, dial, din, dip, discipline,
discuss, disembark, display, dissect, distract, distribute, dive, dock, douse, drag,
draw, drench, drink, drip, drive, drool, drop, drum, dry, duck, dust, dye, eat, educate,
eject, embrace, emerge, empty, encourage, erase, erupt, examine, exercise, exterminate,
extinguish, fall, farm, fasten, feed, fetch, fill, film, fish, fix, flame, flap, flex,
flick, fling, flip, float, floss, fold, forage, ford, frisk, frown, fry, fuel, gamble,
garden, gasp, gather, giggle, give, glare, glow, glue, gnaw, grieve, grill, grimace,
grin, grind, guard, handcuff, hang, harvest, haul, heave, help, hike, hit, hitchhike,
hoe, hoist, hug, hunch, hunt, hurl, ignite, ignore, imitate, immerse, inflate, inject,
insert, instal, instruct, intermingle, interrogate, interview, jog, juggle, jump, kick,
kiss, knead, kneel, knock, lace, land, lap, lather, laugh, launch, lead, leak, lean,
leap, lecture, lick, lift, light, load, lock, make, manicure, march, mash, massage,
measure, mend, microwave, milk, mime, mine, misbehave, moisten, moisturize, mold, mop,
mourn, mow, nag, nail, nip, nuzzle, offer, officiate, open, operate, overflow, pack,
package, paint, panhandle, parachute, parade, paste, pat, paw, pay, pedal, pee, peel,
perform, perspire, phone, photograph, pick, pilot, pin, pinch, pitch, place, plant,
plow, plummet, plunge, poke, poop, pot, pounce, pour, pout, practice, pray, preach,
press, prick, protest, provide, prowl, prune, pry, pucker, pull, pump, punch, punt,
push, put, queue, race, raft, rain, rake, ram, read, rear, reassure, record, recover,
recuperate, rehabilitate, release, repair, rest, restrain, retrieve, rid, rinse, rock,
rot, row, rub, run, salute, say, scold, scoop, score, scrap, scratch, scrub, seal, sell,
serve, sew, shake, sharpen, shave, shear, shell, shelve, shiver, shoot, shop, shout,
shovel, shred, shrug, shush, sign, signal, sing, sit, skate, sketch, ski, skid, skip,
slap, sleep, slice, slide, slip, slither, slouch, smash, smear, smell, smile, sneeze,
sniff, snow, snuggle, soak, soar, socialize, sow, spank, speak, spear, spill, spin,
spit, splash, spoil, spray, spread, sprinkle, sprint, sprout, spy, squeeze, squint,
stack, stampede, stand, staple, star, steer, sting, stir, stitch, stoop, storm, strap,
stretch, strike, strip, stroke, study, stuff, stumble, subdue, submerge, suck, surf,
swarm, sweep, swim, swing, swoop, tackle, talk, tap, taste, tattoo, taxi, teach, tear,
telephone, throw, tickle, tie, till, tilt, tip, tow, train, trim, trip, tug, tune, turn,
twirl, twist, type, uncork, unload, unlock, unpack, unplug, unveil, urinate, vacuum,
vault, videotape, vote, wad, waddle, wag, wait, walk, wash, water, wave, wax, weed,
weep, weigh, weld, wet, wheel, whip, whirl, whisk, whistle, wilt, wink, wipe, work,
wrap, wring, wrinkle, write, yank, yawn
}
\end{footnotesize}

\paragraph{Particles and Prepositions}
\begin{footnotesize}
\textit{
about, after, against, around, as, at, before, behind, by, down, for, from, in, into,
like, of, off, on, onto, out, over, through, to, towards, up, with, without
}
\end{footnotesize}

\subsection{Data Collection}

We provide below the data collection interfaces we used to label properties in both the abstract and situated datasets. The affordances collected for the situated dataset used a similar interface.

\paragraph{Abstract Dataset} The interface for labeling the abstract dataset asks about an object by giving only its name (e.g., \textit{accordion}), and then asking about the \textit{usual} properties of that object. Annotators are given the choice \textit{``too difficult to tell,''} but are encouraged to use that only when absolutely necessary, and use their best guess when possible. We display twenty-five properties at once.

\begin{center}
\includegraphics[width=0.99\columnwidth]{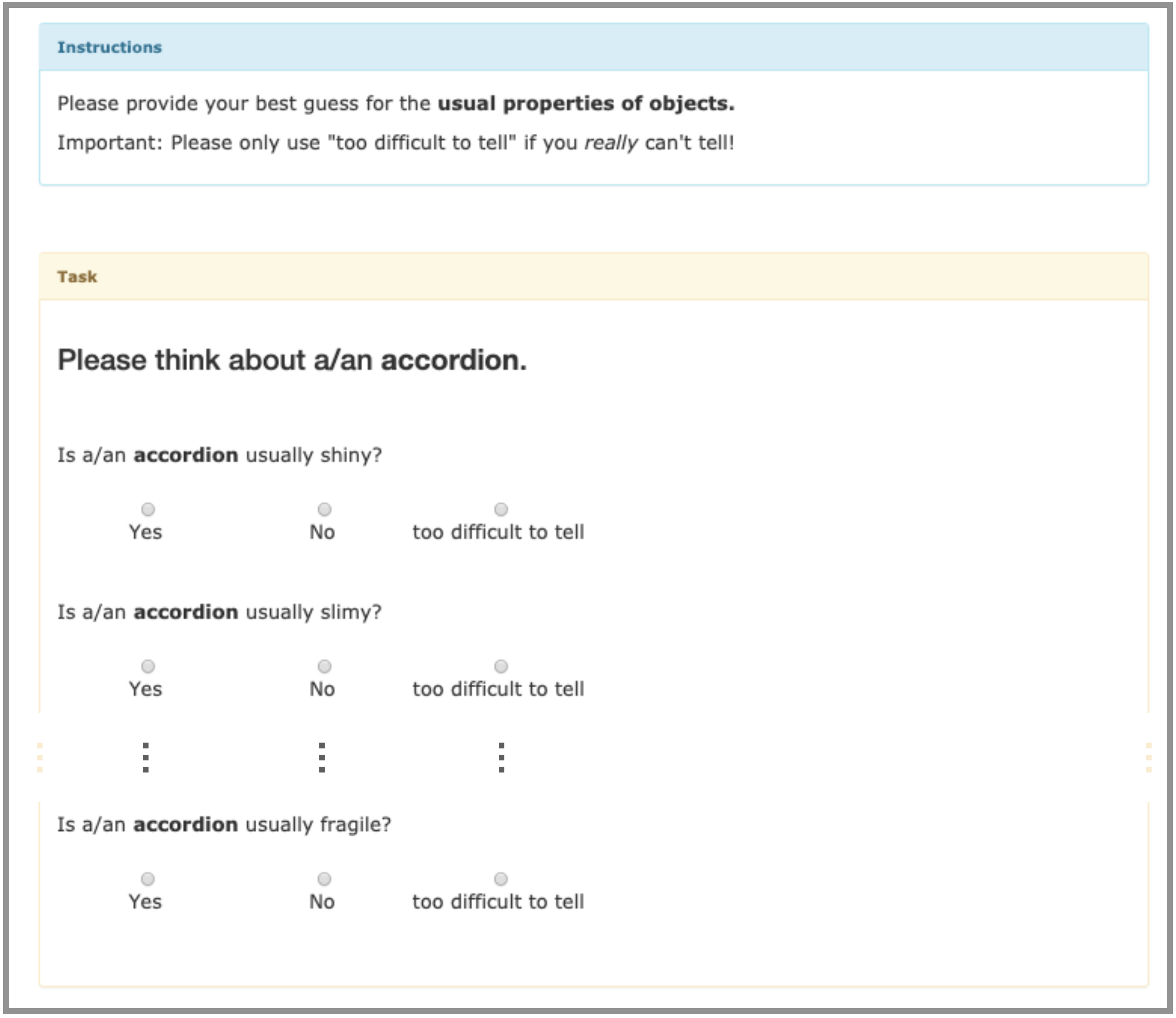}
\end{center}

\paragraph{Situated Dataset} The interface for labeling the situated dataset displays a picture with an object highlighted in it (the photos and object labels are from MS COCO \cite{lin2014microsoft}).

To label properties, the annotator is asked to select which properties apply to this particular object. Because the object is grounded in a specific instance, we remove the \textit{``too difficult to tell''} option, forcing a yes/no decision. We display ten properties at once.

To label \textit{affordances} (not pictured), the annotator is prompted \textit{``What might you do to X?''}, where \textit{X} is the highlighted object. The annotator is asked to provide three to five choices using the provided verbs, particles, and prepositions (described above). After collecting all annotations, we discard the particles and prepositions, and aggregate to pick the top three verbs used for each instance. To create negative samples, we randomly pick three verbs that were not selected from the complete list.

\begin{center}
\includegraphics[width=0.99\columnwidth]{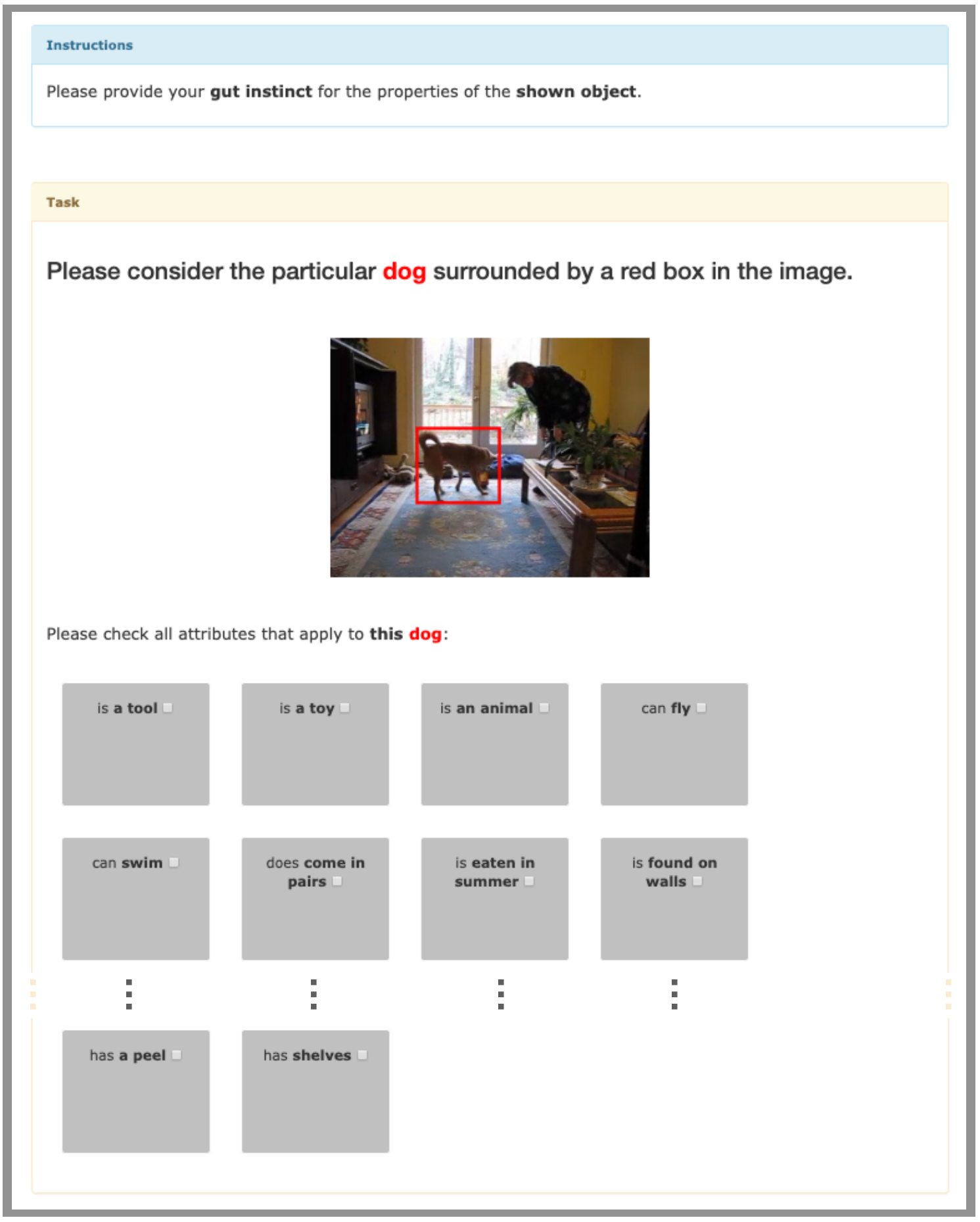}
\end{center}

\paragraph{Quality Control} We collect annotations using Amazon Mechanical Turk. We find workers generally provide high quality annotations. However, even with strict qualification requirements, we often find a nontrivial rate of negligence. To combat this, we inject two pseudo-properties (e.g., \textit{``Is X a word in the English language?''} for the object \textit{X}) in our data collection interface. Because we know the answers to these questions in advance, we can use them as a check to prevent a worker from answering at random. We discard all data from any worker who answers any of these questions incorrectly.

\subsection{Revisions}

\paragraph{August 2019} This is the first version uploaded to arXiv. Previously, BERT was trained in the same way as all other models: fixing the model, and training an MLP on top. We fine-tuned BERT end-to-end, which then outperformed all other models. We updated the \textit{Models} section, results and statistical significance tests (Table~\ref{table:results}), and the analysis graphs (Figure~\ref{fig:results-graphs}). While BERT's finetuned numbers are higher overall, the conclusions of the paper remain unchanged. BERT's performance on the situated affordance $\longleftrightarrow$ property task are still far below humans'.

This version also includes an Appendix, which contains detailed information about our datasets.

\end{document}